\documentclass{article}

    \usepackage[preprint]{neurips_2024}

\usepackage[utf8]{inputenc} 
\usepackage[T1]{fontenc}    
\usepackage{hyperref}       
\usepackage{url}            
\usepackage{booktabs}       
\usepackage{amsfonts}       
\usepackage{nicefrac}       
\usepackage{microtype}      
\usepackage{xcolor}         
\usepackage{graphicx}
\usepackage{soul}
\usepackage{tabularx, booktabs}
\newcolumntype{C}{>{\centering\arraybackslash}X}
\usepackage{amsmath}
\usepackage{amsthm}
\newtheorem{remark}{Remark}
\usepackage{multirow}
\usepackage{makecell}

\title{Targeted Angular Reversal of Weights (TARS) for Knowledge Removal in Large Language Models}

\author{Harry J. Davies\\
  Department of Electrical Engineering\\
  Imperial College London\\
  London, UK \\
  \texttt{harry.davies14@imperial.ac.uk} \\
  \And
  Giorgos Iacovides\\
  Department of Electrical Engineering\\
  Imperial College London\\
  London, UK \\
  \And
  Danilo P. Mandic\\
  Department of Electrical Engineering\\
  Imperial College London\\
  London, UK \\
}

\begin{document}

\maketitle

\begin{abstract}

The sheer scale of data required to train modern large language models (LLMs) poses significant risks, as models are likely to gain knowledge of sensitive topics such as bio-security, as well the ability to replicate copyrighted works. Methods designed to remove such knowledge must do so from all prompt directions, in a multi-lingual capacity and without degrading general model performance. To this end, we introduce the targeted angular reversal (TARS) method of knowledge removal from LLMs. The TARS method firstly leverages the LLM in combination with a detailed prompt to aggregate information about a selected concept in the internal representation space of the LLM. It then refines this approximate concept vector to trigger the concept token with high probability, by perturbing the approximate concept vector with noise and transforming it into token scores with the language model head. The feed-forward weight vectors in the LLM which operate directly on the internal representation space, and have the highest cosine similarity with this refined targeting vector, are then replaced by a reversed targeting vector, thus limiting the ability of the concept to propagate through the model. The modularity of the TARS method allows for a sequential removal of concepts from Llama 3.1 8B, such as the famous literary detective Sherlock Holmes, and the planet Saturn. It is demonstrated that the probability of triggering target concepts can be reduced to 0.00 with as few as 1 TARS edit, whilst simultaneously removing the knowledge bi-directionally. Moreover, knowledge is shown to be removed across all languages despite only being targeted in English. Importantly, TARS has minimal impact on the general model capabilities, as after removing 5 diverse concepts in a modular fashion, there is minimal KL divergence in the next token probabilities of the LLM on large corpora of Wikipedia text (median of 0.002).

\end{abstract}






\section{Introduction} \label{intro}

The ever increasing scale and prominence of large language models (LLMs) comes with a considerable challenge in AI safety. Popular models such as ChatGPT \citep{openai2024gpt4technicalreport} and Llama \citep{llama3herdmodels} are pre-trained to predict the next sequence of characters or words on vast corpora of textual information, encompassing a large proportion of the internet, news articles and digital books. This extensive training enables LLMs to encapsulate a wealth of human knowledge, linguistic patterns, and cultural nuances \citep{eldan2023whosharrypotterapproximate}. However, these benefits come with significant risks, as the training data often includes problematic content, such as copyrighted material \citep{karamolegkou2023copyrightviolationslargelanguage}, toxic or malicious content, misinformation, personal data, and more. \par 
As a result, LLMs are almost certain to learn concepts that can lead to harmful behaviors, raising a variety of privacy, security, and ethical concerns \citep{tofu2024}. For instance, they can generate toxic, offensive, or hateful language \citep{harmful_gen_1,harmful_gen_2} and possess knowledge of sensitive topics, such as bio-security or cyber-security, which could be exploited by malicious actors \citep{sec_1,sec_2}. \par 

\par 
The effectiveness of `knowledge unlearning' algorithms is governed by two main criteria as outlined by \citep{gandikota2024erasingconceptualknowledgelanguage}, that mirror sensitivity and specificity in typical classification paradigms:
\begin{enumerate}
    \item \textbf{Sensitivity (Innocence)}: The knowledge in question should be completely removed, no matter the direction of the prompt. For example, if the model is provided with a description of a harmful category, it should not be able to name the category. Moreover, if asked to describe the category, it should not be able to describe it either. 
    \item \textbf{Specificity}: Knowledge that is not directly related to the removed concept should be maintained, requiring the model editing to be minimally invasive. This can be tested both by exploring how the model talks about closely related topics, and by checking that the general performance of the model is maintained.
\end{enumerate}

In addition to these two core concepts, there is softer requirement that the model output when discussing the harmful topic is not nonsensical \citep{gandikota2024erasingconceptualknowledgelanguage}. We extend these criteria to account for the increasing multilingual capabilities of general-purpose LLMs, highlighting the importance of achieving sensitivity and specificity across \textbf{multiple} languages.

Our method operates based on two underlying assumptions. The first is that knowledge is stored in feedforward networks of large language models \citep{geva2022transformerfeedforwardlayersbuild}, as we explicitly target these layers. The second is that, given the existence of residual connections in LLMs that allow internal representations to bi-pass layers, the high-dimensional representation space itself should be consistent throughout the model. Given these two assumptions, we design a target vector of the same dimension as the model's internal dimension, which contains information that describes the concept we wish to remove. The cosine-similarity is then computed between the target vector and the feedforward network weights which directly operate on this internal representation space, to find weight vectors that have a high affinity to our target vector. In Llama 3.1 8B, the weights we search are the gate-projection and up-projection layers, as these operate directly on the models internal representation of language concepts. Finally, these high-similarity weights are replaced with a reversed targeting vector to ``repel" the knowledge we want to remove. 

The effectiveness of our Targeted Angular Reversal (\textbf{TARS}) method is demonstrated by removing concepts such as of ``Sherlock Holmes" and ``Saturn" from the weights of pre-trained Llama 3.1 8B. To evaluate the sensitivity of our proposed method, it is confirmed that the probability of the token ``Sherlock", when the model is prompted by a detailed description of Sherlock Holmes, is significantly decreased after TARS is applied. Moreover, it is further demonstrated that the edited model is unable to produce a specific description of a Sherlock Holmes, and instead ``hallucinates" descriptions that are either general or factually wrong. In addition, it is shown through the removal of the concept of a ``dog", where different tokens are used in different languages (``chien" in French and ``Hund" in German) that concepts are removed across all languages, even when targeted in English. Importantly, our approach is demonstrated to be \textbf{modular}, meaning that concepts can be removed from the same model sequentially. It is also confirmed that, after modular removal of five such concepts, the posterior distribution of the model is minimally impacted, by measuring the KL-divergence between the base model and the TARS edited model on a large corpus of texts from Wikipedia. \par 
The main contributions of this work are: 
\begin{itemize}
    \item The proposal of \textbf{TARS}, a novel method for knowledge removal that eliminates the need for retraining, significantly reducing the computational overhead typically associated with such techniques.
    \item Our approach is minimally invasive, requiring only a small number of edits to remove any concept. This results in a marginal decrease in the general knowledge capabilities of the TARS-edited model compared to the base model, as demonstrated quantitatively through Kullback–Leibler (KL) divergence between the next-token prediction distributions of the two models. 
    \item To the best of our knowledge, this is the first study to effectively demonstrate both \textbf{non-causal} and \textbf{multilingual} knowledge removal using well-defined criteria, addressing an increasingly important need for knowledge removal, as the multilingual capabilities of LLMs continue to advance.
    \item The effectiveness of our approach is underscored by its \textbf{modularity}, enabling practitioners to precisely control the extent of knowledge removal while preserving the model's general-purpose capabilities. This helps to mitigate the trade-off between knowledge removal and overall model performance observed in existing methods. 
\end{itemize}




\section{Related Work}
A widely adopted approach to safeguard against harmful responses in LLMs is to \textit{align} their generation with safe outputs that follow policy regulations and human values through fine-tuning techniques, particularly Reinforcement Learning from Human Feedback (RLHF) \cite{rlhf_1,rlhf_2}. While effective, this approach is computationally expensive and vulnerable to exploitation by misaligned evaluators. For instance, adversarial prompts can \textit{jailbreak} these fine-tuned models to re-invoke harmful responses \cite{rlhf_lim_1,rlhf_lim_2,rlhf_lim_3}. To address these limitations, an emerging area of research is \textit{knowledge removal}, which focuses on removing and editing specific knowledge associated with undesirable behaviors post-training. Compared to RLHF, knowledge removal is more computationally efficient and easier for practitioners to implement \cite{mu_benefit}. \par 
Many of the knowledge removal methods focus on different ways of fine-tuning the model to remove a particular concept. Representation misdirection for unlearning (RMU) steers the internal representations of knowledge towards a random representation \citep{RMUpaper} which can result in nonsensical outputs. Who is Harry Potter (WHP) trains an initial model to be as knowledgeable as possible about a topic, and then trains the end model to be as different as possible from the first \citep{eldan2023whosharrypotterapproximate}. TOFU shifts the focus from traditional label-specific unlearning to forgetting specific information about individuals in the training data by creating a dataset of 200 synthetic author profiles, each consisting of 20 question-answer (QA) pairs, generated by prompting GPT-4. Since this synthetic data is assumed to be distinct from any existing pretraining data, the unlearning process is unaffected by prior knowledge. However, this comes at the cost of requiring the model to be fine-tuned on the synthetic QA pairs first, to ensure it possesses the targeted knowledge before the unlearning procedure \citep{tofu2024}. The problem with fine-tuning methods is that it is difficult to limit damage to the base model when training to remove a harmful concept, often hindering the general knowledge capabilities of LLMs, while also requiring computational resources to perform in the first place. 
Most importantly, the benchmarks in these works, as well as other commonly cited studies in the knowledge removal community \citep{rwku,tdec,pku_saferlhf}, evaluate unlearning outcomes using a specified set of `forget set' and `retain set' queries. However, as extensively highlighted by \citep{benchmark_1}, forget-retain evaluations are deceptive as real queries likely have dependencies between the forget and retain sets, making such classification challenging, if not impossible. Consequently, minor adaptations in the original queries to enhance their practical application — such as combining questions from both sets or modifying incorrect multiple-choice answers in the retain set to include keywords associated with forget data — have been shown to degrade the performance of all methods significantly \citep{benchmark_1}. \par 
Another possible approach is to locate and modify weights directly without fine-tuning. The Rank-One Model Editing (ROME) method \citep{meng2023locatingeditingfactualassociationsROME} locates and modifies feed forward network weights using causal tracing. Similar to the method we propose in this work, ROME does not require retraining and does not employ the common `forget' and `retain' set evaluation benchmarks.  However, in ROME, the knowledge removal is directional, which means that prompting from a different direction can reveal the knowledge. Moreover, the effectiveness of this approach has not been evaluated across different languages, raising the question of whether the knowledge can be regained by prompting the model in a different language.

\section{Methods}

\begin{figure}[h!]
  \centering
  \includegraphics[width=\textwidth]{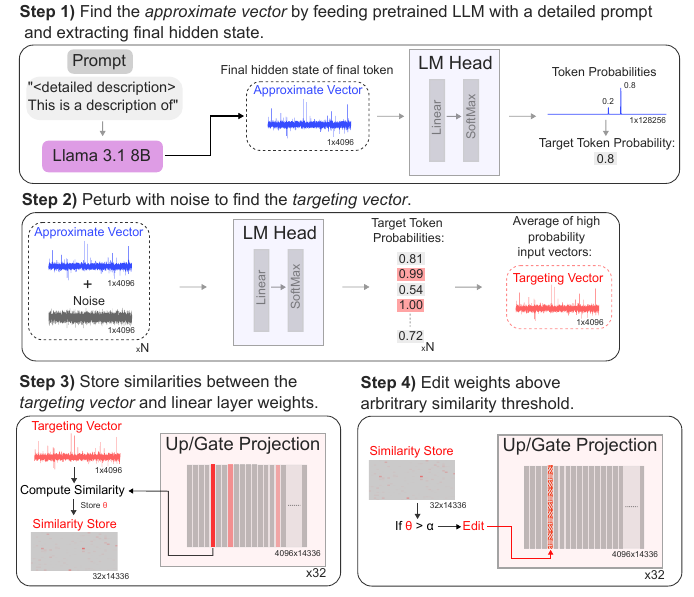}
  \caption{The Targeted Angular Reversal (TARS) method for removing knowledge from large language models. Step 1) The approximate vector (blue) is formed by using the LLM to aggregate information about the concept that is to be removed. Step 2) By creating noisy variants of the approximate vector, a refined Targeting vector (red) that is exclusive to the concept of interest is formed. Step 3) The cosine angle is computed between the targeting vector and all weight vectors in the Up/Gate projection layers. Step 4) Weight vectors above an arbitrary similarity threshold are edited and replaced with a normalised reversed form of the targeting vector.}
    \label{methods}
\end{figure}


Decoder-only transformers \citep{attention_is_all} form the backbone of both the generative pre-trained transformer models, and the Llama models. The decoder-only transformer consists of an attention block, which updates tokens based on the context of preceding tokens, followed by a feedforward block, which can be thought of as computing the results of the attention block. Importantly, there is a linear layer directly after the attention mechanism, which projects the output of attention back to the internal dimension of the model. Without this post-attention projection layer, the internal representation space would not be consistent between the blocks. Decoder-only transformers have residual connections, which allow entire blocks and layers to be skipped, which results in a conservation of this internal representation space across layers. The feed-forward network then acts directly on this internal high-dimensional representation.

Concepts in language are embedded as vectors in this internal representation space. The first layers in each feedforward block can therefore be viewed as a stack of weight vectors where the element-wise product is performed directly with the concept vectors. We can therefore hypothesise that weight vectors in the first layer of each feedforward block, which have a higher affinity with a given concept vector, are then more responsible for the propagation of that concept through the model than weight vectors with a lower affinity. We can measure this affinity or alignment by calculating the cosine similarity between the weight vector and the concept vector. The more aligned a weight vector with a concept, the more likely is the specific element-wise product to sum up to cross an activation threshold. The selective editing of these weight vectors, by replacing them with a reversed form of the concept vector, would result in a negative element-wise product and thus reduce the chance of the concept propagating through the model layers.

This forms the basis of our Targeted Angular Reversal (\textbf{TARS}) method, which can be broken down into four main steps (outlined in Figure.~\ref{methods}):
\begin{enumerate}
    \item Approximating a vector representation for a concept which is to be removed, by using the large language model to build the representation from a detailed descriptive prompt.
    \item Probing the language model head with variants of this approximate vector to refine it into a \textbf{targeting} vector which exclusively triggers the specific concept.
    \item Calculating the cosine \textbf{angle} between weight vectors and the target vector.
    \item Selectively editing weight vectors with high affinity to the targeting vector, by replacing them with a \textbf{reversed} target vector, in order to reduce the likelihood of the concept propagating through the model.
\end{enumerate}

\subsection{Creating a Targeting Vector}


To create the targeting vector, we first leverage the proficiency of large language models for aggregating information in vector space. We collate the relevant descriptive information of a concept into a lengthy description, and complete the prompt with a phrase such as \textit{``This is a description of"}. All relevant information is then aggregated into the last token by the LLM in order to predict the chosen concept with high probability. This is confirmed by transforming the approximation vector into token probabilities via the LM-head, as summarised in Step 1 in Figure.~\ref{methods}. For example, to remove knowledge of the fictional detective Sherlock Holmes from Llama 3.1 8B, our descriptive prompt details his appearance, his relationship with other characters, names of Sherlock Holmes books, and the author Sir Arthur Conan Doyle. The full prompts for this example and others are given in Appendix A1. \par 
Mathematically, let  \( C \) denote a concept, and \( D_C \) its descriptive information. The  prompt is then given by:
\begin{equation}
P_C = D_C + \text{"This is a description of"}
\end{equation} and can be tokenized into a sequence of tokens:
\begin{equation}
[t_1, t_2, \ldots, t_n],
\end{equation}
where \( t_n \) is the final token.

If \( f_{\text{LLM}} \) is the LLM's encoding function, then the hidden states \( \mathbf{h_1}, \mathbf{h_2}, \ldots, \mathbf{h_n} \) are computed as
\begin{equation}
\mathbf{h_i} = f_{\text{LLM}}(t_1, t_2, \ldots, t_i), \quad i = 1, \ldots, n.
\end{equation} The aggregated representation at the final token is then
\begin{equation}
\mathbf{h_n} = f_{\text{LLM}}(t_1, t_2, \ldots, t_n),
\end{equation} and the model's probability distribution over the vocabulary is given by
\begin{equation}
p(\mathbf{v_{\text{approx}}} \mid P_C) = \text{SoftMax}(\mathbf{W_{\text{head}} h_n + b_{\text{head}}}),
\end{equation} where \( \mathbf{W_{\text{head}}} \) and \( \mathbf{b_{\text{head}}} \) denote the parameters of the LM head and $\mathbf{v_{\text{approx}}}$ is the approximate concept vector. The target token probability is then equal to
\begin{equation}
p_{\text{target}} = \max_{\mathbf{v} \in \text{Vocabulary}} p(\mathbf{v_{\text{approx}}} \mid P_C).
\end{equation}
\\
\begin{remark}
Much of the effectiveness of TARS is derived from the richness of information included in the prompt which is used to generate the approximation vector. Another simple way to trigger the token `` Sherlock" with high probability would be to use the prompt ``An eight letter word with the combination of the letters S, h, e, r, l, o, c and k is the word". However, this produces an internal representation with comparatively less information about Sherlock Holmes, meaning that this information cannot be sought out and penalised in the model.
\end{remark}

A potential issue with this information-rich approximation vector is that it could also align with other related concepts, and may contain information that is not contributing to the LLMs determination of the concept. To circumvent this issue, we refine the approximation vector to produce a vector which triggers the concept with the probability of 95\% or higher. The refinement is performed by firstly adding noise to the approximation vector across all dimensions. Noisy approximation vectors are then stored if they trigger the resulting probability of the concept as 0.95 or greater. This process is repeated many times (in the case of \textit{"Sherlock"}, a batch size of 450 was repeated 25,000 times to create a pool of candidate vectors). An average of the resulting candidate vectors is then taken to produce a \textbf{targeting} vector, as described in Step 2 in Figure~\ref{methods}. \par 
Mathematically, define the noisy vector \( v_{\text{noise}}^{(i)} \) as
\begin{equation}
\mathbf{v_{\text{noise}}^{(i)}} = \mathbf{v_{\text{approx}}} + \mathbf{\epsilon^{(i)}},
\end{equation}
where \( \mathbf{\epsilon^{(i)}} \sim \mathcal{N}(0, \sigma^2 I) \) represents a vector of Gaussian noise with standard deviation \( \sigma \) (e.g., $\sigma=10$).

For each \( \mathbf{v_{\text{noise}}^{(i)}} \), compute the probability \( p_i \) of the concept \( C \) being triggered
\begin{equation}
p_i = p(t_C \mid \mathbf{v_{\text{noise}}^{(i)}}),
\end{equation}
where \( t_C \) denotes the token representing the concept \( C \), and \( p(t_C \mid \mathbf{v_{\text{noise}}^{(i)}}) \) is the probability from the LLM's SoftMax distribution.

Next, a candidate vector \( \mathbf{v_{\text{noise}}^{(i)}} \) is then retained if
\begin{equation}
p_i \geq \tau,
\end{equation}
where \( \tau \) is a predefined threshold (e.g., \( \tau = 0.95 \)) resulting in a  set of retained candidate vectors
\begin{equation}
\mathbf{V_{\text{cand}}} = \{ \mathbf{v_{\text{noise}}^{(i)}} : p_i \geq \tau \}.
\end{equation}

The \textit{targeting vector} \( \mathbf{v_{\text{target}}} \) is then computed as the mean of the retained candidates
\begin{equation}
\mathbf{v_{\text{target}}} = \frac{1}{N} \sum_{\mathbf{v} \in \mathbf{V_{\text{cand}}}}\mathbf{v}
\end{equation}
where \( N = |\mathbf{V_{\text{candidates}}}| \) denotes the number of retained candidate vectors. \\

\begin{remark}
Whilst this step seems computationally intense, the LM head is a simple network consisting of a single linear layer and a SoftMax output. The combination of large batches to parallelize the computation and the simple network means that this experimentation to refine the approximation vector can be performed fast. 
\end{remark}

\subsection{Locating Knowledge Weights}

In Llama 3.1, the feed-forward network consists of a gated linear unit \citep{shazeer2020gluvariantsimprovetransformer}, and there are therefore two sets of weights to search in each layer, namely the "up-projection" weights and the "gate-projection" weights. In Llama 3.1 8B, the internal model dimension is 4,096, which is the dimension of our targeting vector. The linear layer matrix is a collection of 4,096 length vectors, of which the dot product is computed with the internal model dimension during the forward pass. As discussed previously, we operate on these weights with the assumption that the internal representation space is preserved throughout the model due to the presence of residual connections.

To locate the weight vectors which have the highest affinity to our targeting vector, \(\mathbf{ v_{\text{target}}} \), we calculate the angle (cosine similarity) between each weight vector and our targeting vector. The Llama 3.1 8B model has gate-projection matrices, \( \mathbf{W_{\text{gate}}^{(\ell)}} \), and up-projection matrices, \( \mathbf{W_{\text{up}}^{(\ell)}} \), of dimension  $4096 \times 14336$, in each of the 32 layers. For each layer, we compute the cosine similarity for all \( 14,336 \) weight vectors, \( \mathbf{w^{(\ell)}_i} \), in both \( \mathbf{W_{\text{gate}}^{(\ell)}} \) and \( \mathbf{W_{\text{up}}^{(\ell)}} \), as 
\begin{equation}
S_C(\mathbf{w^{(\ell)}_i}, \mathbf{v_{\text{target}}}) = \frac{\mathbf{w^{(\ell)}_i} \cdot \mathbf{v_{\text{target}}}}{\|\mathbf{w^{(\ell)}_i}\| \|\mathbf{v_{\text{target}}}\|}
\end{equation}
resulting in a total of $14,336 \times 32 \times 2$ cosine similarity scores, providing similarities for 917,504 candidate vectors for potential modification.  This is step 3 in Figure~\ref{methods}.\\
\begin{remark}
To achieve an effective knowledge removal of \textit{`Sherlock'}, we only edit 1 (0.0001\%) of these 917,504 weight vectors, to reduce their alignment with \( v_{\text{target}} \).
\end{remark}

\subsection{Editing Strategies}

The 4th and final necessary step for TARS is to replace candidate weights with the reverse of the target vector. This reversed target vector is also normalised with the 3-norm. Candidate weights are determined as weight vectors with a cosine-similarity above an arbitrary threshold, $\theta$. Formally, for each weight vector \( \mathbf{w^{(\ell)}_i} \in \mathbf{W_{\text{candidates}}} \), replace
\begin{equation}
\mathbf{w^{(\ell)}_i} \leftarrow  - \frac{\mathbf{v_{\text{target}}} }{\| \mathbf{v_{\text{target}}} \|_3} ,
\end{equation}
where
\[
\mathbf{W_{\text{candidates}}} = \{ \mathbf{w^{(\ell)}_i} : S_C(\mathbf{w^{(\ell)}_i}, \mathbf{v_{\text{target}}}) > \theta \}. 
\] \par 
In practice, we suggest gradually lowering this threshold to edit more target vectors, and examining performance to determine the appropriate threshold. To reduce the likelihood of the concept from propagating through the model when it arises, we replace these weight vectors that have the strongest response to our targeting vector, with the reversed targeting vector, as in Equation 13, illustrated step 4 in Figure~\ref{methods}. There is scope for optimization of the amplitude of the reversed target vector to further improve the sensitivity of TARS, that was not explored in this work.

\section{Results}

\begin{figure}[h!]
  \centering
  \includegraphics[width=\textwidth]{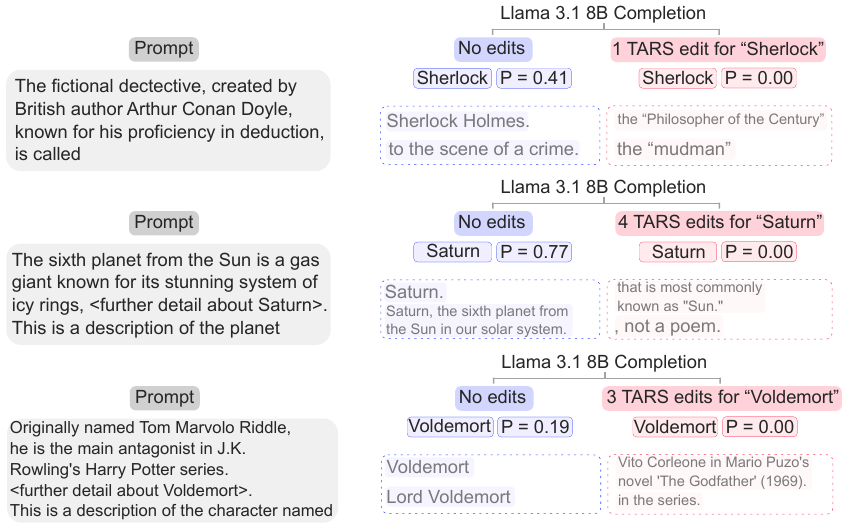}
  \caption{The probability (P, right) that the next token generated by the LLM is the targeted concept, given the input prompt (grey, left), along with two example responses, for Llama 3.1 8B with (red) and without (blue) TARS edits. Top) Targeted removal of knowledge relating to fictional detective ``Sherlock Holmes". Middle) Targeted removal of knowledge relating to the planet Saturn. Bottom) Targeted removal of knowledge relating to the Harry Potter villain ``Voldermort". }
    \label{sherlock_proba}
\end{figure}

The first step to quantify knowledge removal post TARS for different concepts is to check the change in probability of the target token being next token predicted by the LLM, given a description of the target in the input prompt. Observe from Figure~\ref{sherlock_proba} that after only 1 TARS edit to remove ``Sherlock", the probability of the token ``Sherlock" given the input prompt is reduced from 0.41 to 0.00. Similarly, after 4 edits for ``Saturn" the probability of Saturn is reduced from 0.77 to 0.00, and after 3 edits for ``Voldemort" the probability of ``Voldemort" is reduced from 0.19 to 0.00. The model instead completes the prompt with either general responses, such as ``the Philosopher of the Century" in the case of Sherlock Holmes, or with factually wrong responses such as ``Vito Corleone" in the case of ``Voldemort" and ``Sun" in the case of Saturn. Each of these example completions are shown below the probabilities for each concept in Figure~\ref{sherlock_proba}. Importantly, responses from the model are still somewhat related to the prompt, rather than just outputting random characters.

\subsection{The Bi-directionality of Knowledge Loss}


\begin{figure}[h!]
  \centering
  \includegraphics[width=\textwidth]{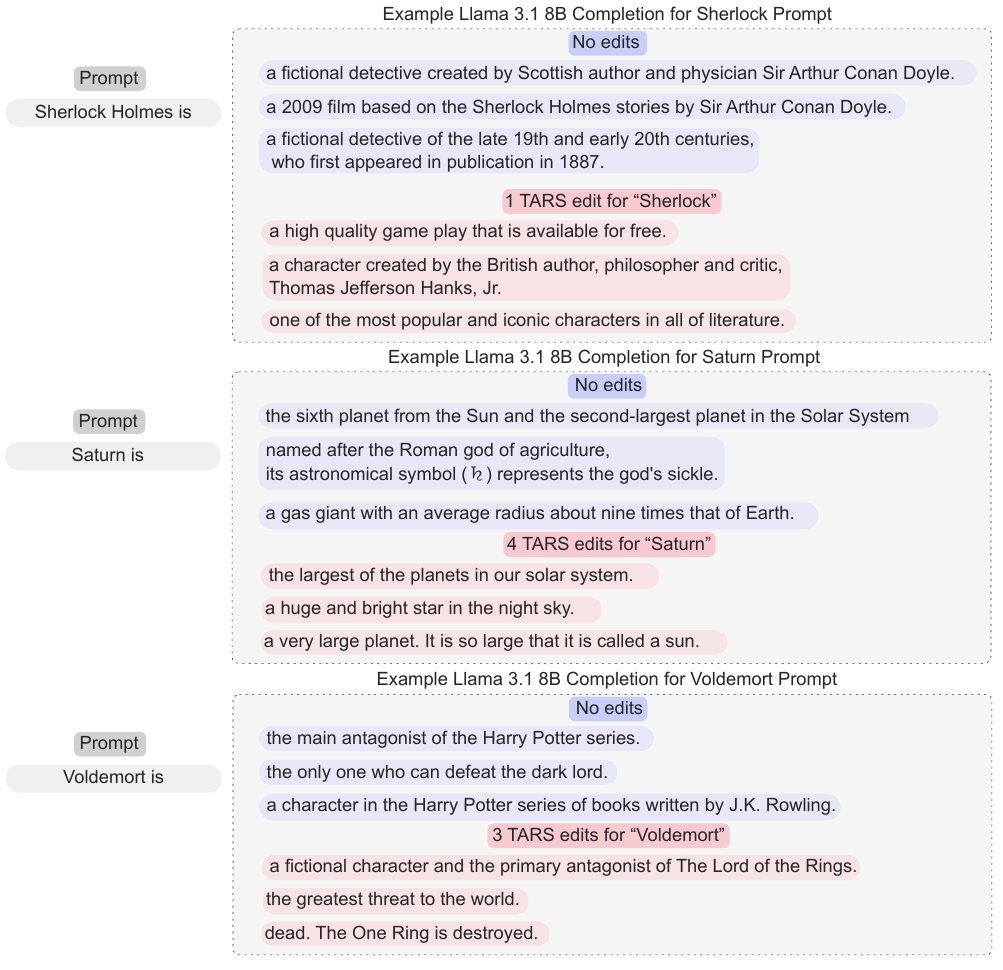}
  \caption{Evaluation of non-causal knowledge loss via TARS removal, with input prompts which ask for a description of each removed concept (grey, left). For each example, three model completions are provided for Llama 3.1 8B without edits (blue) and with TARS edits (red). Top) Targeted removal of knowledge relating to fictional detective ``Sherlock Holmes". Middle) Targeted removal of knowledge relating to the planet Saturn. Bottom) Targeted removal of knowledge relating to the Harry Potter villain ``Voldemort".}
    \label{bidirectional_all}
\end{figure}

\begin{remark}
As stated Section \ref{intro}, it is not enough to only remove the concept in a causal fashion. Useful knowledge removal requires that the model should not only be unable to classify the concept accurately based on description, but also that the model should not be able to produce an accurate description when prompted to describe the concept.
\end{remark}

To examine this, we prompted the pre-trained model with prompts such as \textit{``Sherlock Holmes is"} and \textit{``Saturn is"}, both before and after the application of TARS to remove the concept of ``Sherlock" and ``Saturn". For each application of TARS, in Figure~\ref{bidirectional_all} we show three example LLM completions, both before the edits and after the knowledge removal edits. It is demonstrated in the case of Sherlock that, before our edits, the model produces a succinct and accurate description of Sherlock Holmes, referencing Sir Arthur Conan Doyle, as well as the period in which the books were set. After only 1 edit with TARS, the model responds generally by stating that Sherlock Holmes is ``a high quality game" and also falsely by stating that it Sherlock Holmes was written by ``Thomas Jeffserson Hanks". Similarly, for Saturn, before editing the model is able to correctly recall facts such as ``the sixth planet from the Sun" before TARS, and after 4 edits with TARS it states falsities such as Saturn is ``the largest of the planets in our solar system". This pattern is repeated again with ``Voldemort", where the un-edited model correctly states that ``Voldemort" is ``the main antagonist of the Harry Potter series.", and after 3 edits with TARS the model instead associates ``Voldemort" with ``The Lord of the Rings". It is evidenced by these results that knowledge removal via TARS is non-causal, as it specifically targets the models internal representation of a concept.

\subsection{Generalisation of Knowledge Loss Across Languages}


\begin{figure}[h!]
  \centering
  \includegraphics[width=\textwidth]{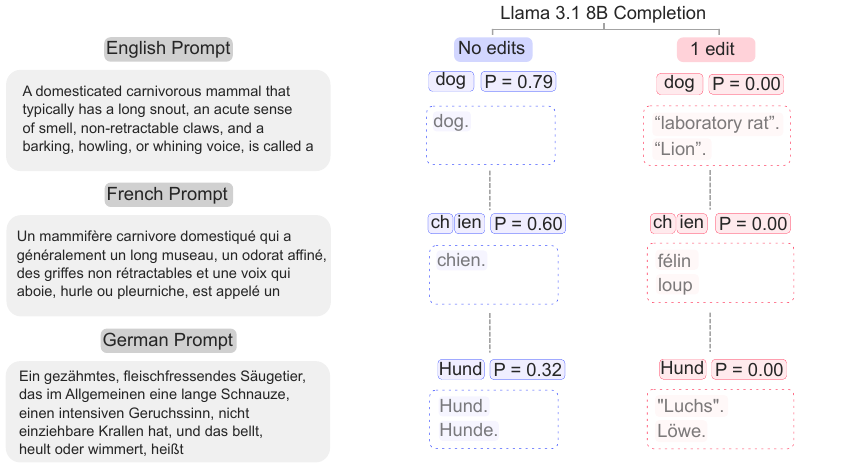}
  \caption{Evaluation of the results of removal of the concept ``dog", with the TARS method applied in English, across the other languages of French and German. The probability (P, right) that the next token generated by the LLM is the targeted concept in differnt languages, given the input prompts in different languages (grey, left), along with example responses, for Llama 3.1 8B with (red) and without (blue) TARS edits. The TARS method is only ever applied in English in this example. Top) The prompt, and the corresponding probability and example completions for ``dog" in English. Middle) The prompt, and the corresponding probability and example completions for ``chien", which is ``dog" in French. Bottom) The prompt, and corresponding probability and example completions for ``Hund", which is ``dog" in German.}
    \label{main_fg}
\end{figure}

For practical application in real-world multilingual LLMs, knowledge removal should be consistent across all languages. Otherwise, a user could simply prompt the LLM in a different language to retrieve the harmful knowledge. To this end, we translated our English description prompt for ``dog" in different languages, with assistance from native French and German speakers. The concept ``dog" was chosen specifically given that it has distinct tokens in different languages, vs a concept such as ``Sherlock" which would be the same token in different languages. It is demonstrated in Figure~\ref{main_fg} that the TARS method removes the concept dog in both French and German, despite ``dog" being targeted for removal in English. With the descriptive prompt designed to trigger the model to complete with the concept of dog, the probability of the token ``dog" is reduced to 0.00 after 1 edit. Importantly, it is shown that the probability of ``chien" after a prompt describing dog in French is also reduced to 0.00, and similarity the probability of ``Hund" is reduced to 0.00 after a German prompt. The example completions in English switch from ``dog" to ``Lion" and ``laboratory rat". In French, completions change from ``chien" (dog) to ``félin" (meaning a member of the cat family, such as lions or cheetahs) and ``loup" (meaning wolf). In German, completions change from ``Hund" (dog) to ``Luchs" (meaning lynx) and ``Löwe" (meaning lion).

\begin{remark}
The results in Figure~\ref{main_fg} are as are expected, as whilst we target the models internal representation of the concept in English, recent evidence by Wendler \textit{et al.} \citep{wendler2024llamasworkenglishlatent} suggests that Llama models use an English internal representation, before translating back to the target language.
\end{remark}

\begin{figure}[h!]
  \centering
  \includegraphics[width=\textwidth]{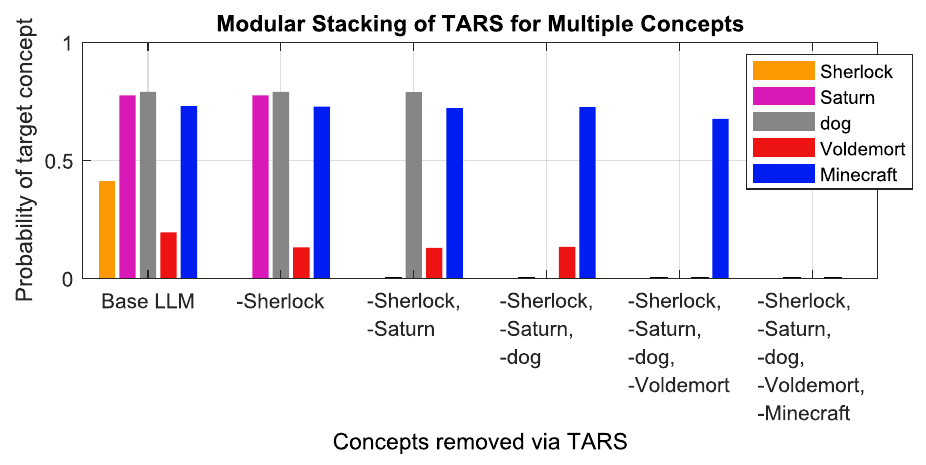}
  \caption{Demonstration of the modular capabilities of TARS. Target token probabilities are extracted from the Llama 3.1 8B, with no edits, and after each concept sequentially removed from left to right. The probability is represented by bars, for the concepts of ``Sherlock" (orange), ``Saturn" (pink), ``dog" (grey), ``Voldemort" (red) and ``Minecraft" (blue).}
    \label{modular_all}
\end{figure}

\subsection{Modular knowledge removal}

The TARS method is modular, meaning that practitioners can stack multiple concepts for removal. This allows for a very limited degradation in the underlying performance of the model, as model performance can easily be evaluated after each concept is removed. Importantly, if a practitioner deploys the LLM, but decides that removal of another concept is necessary for either safety or copyright reasons, it is simple to just remove one more concept without needing to repeat the process for all concepts again. The modularity of TARS is evidenced in Figure.~\ref{modular_all} which shows the sequential degradation of the target probabilities as TARS is sequentially applied to several concepts.

\subsection{Maintenance of General Model Performance}

\begin{table}[h]
\centering 
\resizebox{\textwidth}{!}{%
\begin{tabular}{ccc}
 & \bf Text Datasets & \bf \makecell{KL Divergence of TARS-Edited Llama \\ (Median [90\% CI])} \\  \midrule
\multirow{2}{*}{\bf General Knowledge} & Wikitext 2 (Test) & $\mathbf{0.0015} \: [0.0009-0.0049]$ \\
& Wikitext 2 (Training) & $\mathbf{0.0015} \: [0.0008-0.0075]$ \\
\midrule
\multirow{5}{*}{\bf Removed Knowledge} & Sherlock Wikipedia Page & $\mathbf{0.8867} \: [0.2694-2.6967]$ \\
& Saturn Wikipedia Page & $\mathbf{0.1733} \: [0.0867-0.4644]$ \\
& Dog Wikipedia Page & $\mathbf{0.0915} \: [0.0199-0.3192]$ \\
& Voldemort Wikipedia Page & $\mathbf{0.5220} \: [0.1917-0.8761]$ \\
& Minecraft Wikipedia Page & $\mathbf{0.3933} \: [0.0822-0.6730]$ \\
\bottomrule
\end{tabular}
}
\caption{The Kullback–Leibler (KL) divergence between the posterior predicted next token probabilities of Llama 3.1 8B without knowledge removal, and Llama 3.18B after 5 concepts are removed via TARS. The KL divergence is calculated across general datasets of text from Wikipedia \citep{merity2016pointer}, as well as Wikipedia pages corresponding to the specific concepts that were targeted for removal via TARS.}
\label{table:KL}
\end{table}

The utility of knowledge removal is governed by its specificity. It is essential that the removal of specific concepts does not degrade the general performance  of the model. To examine the performance of TARS, we compute the Kullback–Leibler (KL) divergence between the posterior predicted next token probabilities of Llama 3.1 8B without knowledge removal, and Llama 3.1 8B after 5 concepts are removed via TARS. The KL divergence was computed on a large corpus of Wikipedia text \citep{merity2016pointer} to assess the maintenance of general model performance, and compared with the KL divergence on Wikipedia pages of the concepts that were targeted for removal. Table~\ref{table:KL} shows that the median KL divergence of TARS edited Llama on both the Wikitext 2 training and test sets is 0.0015, indicating minimal changes to general capabilities of the underlying model. Conversely, when KL divergence was examined on Wikipedia pages corresponding to a targeted concept, median KL divergence ranged from 0.0915 for ``dog" to 0.8867 for ``Sherlock" (approximately 60-fold to 590-fold higher than divergence on general knowledge).









\section{Conclusions}
We have introduced a novel approach for knowledge removal that rests on reversing the embedding vector of the targeted concept. Our method is computationally efficient, as it eliminates the need for retraining, which is typically required by most knowledge removal techniques. Importantly, it is minimally invasive, as it is capable of removing any concept with very few edits, thereby preserving the general knowledge capabilities of the model. To the best of our knowledge, this study is the first to demonstrate bi-directional and multilingual knowledge removal without the need of fine-tuning, criteria that are essential for effective and practical knowledge removal. Finally, the modular nature of our approach offers practitioners greater flexibility to sequentially remove knowledge from the model, allowing precise control over the trade-off between model utility and knowledge unlearning.

\bibliographystyle{unsrtnat}
\bibliography{IEEEabrv,safellama}

\appendix


\section{Appendix}

\subsection{Prompts to form the approximate vectors}

\subsubsection{Sherlock Holmes}

``Imagine a brilliant detective with an unparalleled knack for observation and deduction, often seen donning a deerstalker hat and an Inverness cape, and possessing a razor-sharp intellect capable of piecing together the most obscure clues to solve complex mysteries. This character's adventures, frequently chronicled by Dr. John Watson, have been adapted into numerous TV series, including the modern 'Sherlock' starring Benedict Cumberbatch, and the classic Granada Television series with Jeremy Brett. The detective's cinematic portrayals include Robert Downey Jr.'s action-packed films and Ian McKellen's reflective 'Mr. Holmes.' Originating from the novels and short stories by Sir Arthur Conan Doyle, such as 'A Study in Scarlet' and 'The Hound of the Baskervilles,' this iconic figure has become a cultural touchstone, inspiring countless adaptations and new works. Their analytical mind, complemented by a deep understanding of human nature, makes them a formidable investigator and a keen observer of society, with a loyal friend in Watson, a nemesis in Professor Moriarty, and a comforting presence in Mrs. Hudson at their Baker Street residence. This is a description of"

\subsubsection{Saturn}

``The sixth planet from the Sun is a gas giant known for its stunning system of icy rings, which are the most extensive and complex in the solar system. This planet is primarily composed of hydrogen and helium, making it less dense than water, so it would float if placed in a sufficiently large body of water. It has a diameter about ten times that of Earth, making it the second-largest planet in our solar system. This celestial body has been observed since ancient times and was the farthest planet visible to the naked eye before the invention of the telescope. It takes approximately 29.5 Earth years to complete one orbit around the Sun. The planet's atmosphere is characterized by strong winds and storms, including the Great White Spot, a massive storm that appears roughly every 30 years. It has 146 known moons, with Titan being the largest, even bigger than the planet Mercury. Titan is unique for its dense atmosphere and liquid lakes of methane and ethane. The planet's exploration has been significantly advanced by missions such as Pioneer 11, Voyager 1 and 2, and the Cassini-Huygens mission, which provided detailed images and data about its rings, moons, and atmospheric condition. This is a description of the planet"

\subsubsection{dog}

``Imagine a loyal and affectionate companion, known for its keen sense of smell and hearing, often seen wagging its tail in excitement. This four-legged friend, belonging to the genus Canis and scientifically named Canis familiaris, comes in a variety of breeds, each with unique characteristics, from the tiny, energetic Chihuahua to the large, gentle Great Dane. With a coat that can range from short and sleek to long and fluffy, this animal is a beloved member of many households. It thrives on companionship and enjoys activities like fetching a ball, going for walks, and playing in the park. Known for its intelligence and trainability, this creature can learn a wide array of commands and tricks, making it a favorite in obedience and agility competitions. Its expressive eyes and ability to sense human emotions make it an excellent therapy animal, providing comfort and support to those in need. Whether serving as a guide for the visually impaired, a member of a search and rescue team, or simply a cherished pet, this animal's unwavering loyalty and joyful spirit make it a treasured part of human life. The history of this companion dates back around 15,000 years, when it was first domesticated from wolves by early humans. Initially serving as hunting partners and protectors, these animals evolved alongside humans, adapting to various roles and environments, from sled dogs in Siberia to sacred animals in ancient Egypt. They are naturally protective, often guarding their owners and territory due to their pack instincts. Their diet has evolved from their carnivorous ancestors. They have non-retractable claws, which provide traction and stability while running. This is a description of a"

\subsubsection{Voldemort}

``Originally named Tom Marvolo Riddle, he is the main antagonist in J.K. Rowling's Harry Potter series. First appearing in 'Harry Potter and the Philosopher's Stone' (1997), he is the archenemy of Harry Potter, who, according to a prophecy, has 'the power to vanquish the Dark Lord.' After murdering Harry's parents, this character attempts to kill Harry, leaving him with a lightning bolt-shaped scar. Feared by nearly every witch and wizard, he is often referred to as 'You-Know-Who' or 'He-Who-Must-Not-Be-Named.' His obsession with blood purity drives his aim to eliminate Muggle heritage and dominate both the Muggle and wizarding worlds. As the last descendant of Salazar Slytherin, he leads the Death Eaters in his quest for power. Throughout the series, his name is so feared that it becomes taboo, allowing his followers to trace anyone who speaks it. His name, derived from French, means 'flight of death' or 'theft of death.' This is a description of "

\subsubsection{Minecraft}

``The game we are talking about is a popular sandbox game developed by Mojang Studios and released in 2011 that was originally created by Markus "Notch" Persson using Java. After its full release, Jens "Jeb" Bergensten took over development. Players can build and explore virtual worlds made up of blocks. The game, which Microsoft acquired in 2014 for \$2.5 billion, has become the best-selling video game ever, with over 300 million copies sold and nearly 170 million monthly active players as of 2024. Players explore a procedurally generated, voxel-based world, gathering resources, crafting items, and building structures. It features multiple game modes, including survival and creative, and supports multiplayer interactions. The game we are talking about is named"

\end{document}